\pdfoutput=1
\documentclass[11pt]{article}

\usepackage[]{emnlp2022}

\usepackage{times}
\usepackage{balance}
\usepackage{latexsym}
\usepackage{tabularx}
\usepackage{enumitem}

\usepackage[T1]{fontenc}
\usepackage[utf8]{inputenc}

\usepackage{microtype}
\usepackage[utf8]{inputenc} % allow utf-8 input
\usepackage[T1]{fontenc}    % use 8-bit T1 fonts
\usepackage{hyperref}       % hyperlinks
\usepackage{url}            % simple URL typesetting
\usepackage{booktabs}       % professional-quality tables
\usepackage{amsfonts}       % blackboard math symbols
\usepackage{nicefrac}       % compact symbols for 1/2, etc.
\usepackage{microtype}      % microtypography
\usepackage{xcolor}         % colors
\usepackage{graphics}    
\usepackage{balance}
\usepackage{latexsym}
\usepackage{tabularx}
\usepackage{enumitem}
\usepackage{tabularray}
\usepackage{amsmath}
\usepackage[T1]{fontenc}
\usepackage{bbm}
\usepackage[utf8]{inputenc}
\usepackage{microtype}
\usepackage[utf8]{inputenc} % allow utf-8 input
\usepackage[T1]{fontenc}    % use 8-bit T1 fonts
\usepackage{hyperref}       % hyperlinks
\usepackage{url}            % simple URL typesetting
\usepackage{booktabs}       % professional-quality tables
\usepackage{amsfonts}       % blackboard math symbols
\usepackage{nicefrac}       % compact symbols for 1/2, etc.
\usepackage{microtype}
\usepackage{graphicx}
\usepackage{graphics}    
\usepackage{tabularx}
\definecolor{mygreen}{RGB}{48, 110, 105}
\definecolor{myblue}{RGB}{30, 76, 124}
\definecolor{mygray}{RGB}{94, 94, 94}
\definecolor{mylightblue}{RGB}{62, 128, 187}
\definecolor{myblack}{RGB}{0, 0, 0}
\title{Comparing Apples to Apples: Generating Aspect-Aware Comparative Sentences from User Reviews}
\author{  Jessica Maria Echterhoff \\
  UC, San Diego\\
  La Jolla, CA 92037 \\
  \texttt{jechterh@ucsd.edu} \\
  \And 
  An Yan \\
  UC San Diego\\
  La Jolla, CA 92037 \\
  \texttt{ayan@ucsd.edu} \\
   \And
  Julian McAuley \\
  UC San Diego \\
  La Jolla, CA 92037 \\
  \texttt{jmcauley@ucsd.edu} }

\begin{document}
\maketitle
\begin{abstract}
It is time-consuming to find the best product among many similar alternatives. \emph{Comparative sentences} can help to contrast one item from others in a way that highlights important features of an item that stand out. Given reviews of one or multiple items and relevant item features, we generate comparative review sentences to aid users to find the best fit. Specifically, our model consists of three successive components in a transformer: (i) an item encoding module to encode an item for comparison, (ii) a comparison generation module that generates comparative sentences in an autoregressive manner, (iii) a novel decoding method for user personalization. We show that our pipeline generates fluent and diverse comparative sentences. 
We run experiments on the relevance and fidelity of our generated sentences in a human evaluation study and find that our algorithm creates comparative review sentences that are relevant and truthful. 
%\footnote{The code and datasets used in this study are available at \url{https://github.com/anonymous/}}
\end{abstract}

\section{Introduction}
Deciding on a product to purchase can be a time-consuming process. Every user has specific quality preferences, budget restrictions, or enjoys different item features. To distill important information, users typically have to read the specifications and reviews of several different (but often very similar) products. For example, when a user is interested in digital pianos, they might want to know which item has the best price-to-value ratio, longevity, or sound quality. If the sound quality of an electronic piano is of importance for one user, the sentence \emph{``This piano sounds more natural than my Sony NWZ-A855."} can give richer information than \emph{``This piano sounds natural."}. 

\begin{table}[t]
    \centering
    \begin{tabularx}{\linewidth}{X}
        Musical Instruments Data \\
        \midrule
        \color{mylightblue} \textbf{Extractive justification:} the \underline{sound} is very good: warm, surprisingly loud, very pleasant to the ear.\newline
        \color{myblue} \textbf{Extractive comparison:}The intonation is spot on (I compared it to my Conn trumpet).\newline
        \color{teal} \textbf{Abstractive comparison PPLM:} Even with an acceptable tone that seems to be nice and bright for your liking (fuller than any other cheap Chinese horn)\newline
        \color{mygreen} \textbf{Abstractive comparison T5 (ours):} am not sure what is, but I find the \underline{valve} to be smoother without any strain.\newline
        \color{mygray} \textbf{Abstractive comparison T5-AGG (ours):} is a very good pocket trumpet, I have to say that it \underline{sound}s better than the cheap Chinese ones.
        \color{myblack}\\
        \midrule
    \end{tabularx}
    \begin{tabularx}{\linewidth}{X}
        Electronics Data \\
        \midrule
         \color{mylightblue} \textbf{Extractive justification:} A good value for your money.\newline
        \color{myblue} \textbf{Extractive comparison:} These are the \underline{earbuds} I tend to prefer above the more expensive ones.\newline
        \color{teal} \textbf{Abstractive comparison PPLM:} has better \underline{sound quality} than that one.\newline
        \color{mygreen} \textbf{Abstractive comparison T5 (ours):} probably the best value for money in my house.\newline
        \color{mygray} \textbf{Abstractive comparison T5-AGG (ours):} \underline{sound quality} is better than those white \underline{iPod earphones.}\newline
        \color{myblack}\\
        \midrule
    \end{tabularx}
    \caption{Generation Results from this study. We use an abstractive approach to generate aspect-aware comparative sentences that underline \underline{product aspects}.}
    \label{tab:examples}
\end{table}
We define a comparative sentence as a \emph{sentence highlighting differences of an item to ground its relative perception}. A comparative sentence typically includes a comparative adjective or adverb (such as ``bigger", ``faster", ``more beautiful", ``less expensive", etc.). Comparative sentences assist users in their relative perception of items and can provide a richer information context to point out specific features that are superior or inferior compared to other similar products. Recommendation explanations like justifications~\citep{ni2019justifying} or `tips'~\citep{li2019persona} assess qualities of a single product and are evaluative by nature, but do not put the generated explanation into a larger context. There are few prior works that tangentially study the problem of generating indirect comparisons~\citep{yang2021comparative, ginty2002comparison, 10.1145/3477495.3532065}, but we are the first to directly generate \emph{abstractive} recommendations for this task, which means that rather than relying on templates or \emph{extractions} from previously written reviews, our method has the power of abstracting from previous reviews to generate new, personalized comparative recommendation explanations. \\
\textbf{Our work} generates sentences that express relative comparisons to facilitate product purchase decisions. Our method can additionally underline outstanding product features for user-specific personalization. In particular, we make the following contributions: 
\begin{enumerate}[left=0pt]
\item  We provide a BERT-based classifier to extract comparative sentences from large review corpora to train generative language models.  Using this classifier, we provide a comparative sentence dataset of $258,816$ extracted comparative sentence instances, which is $35\times$ as large as previously generated comparative sentence datasets~\citep{franzek2018categorization, panchenko2018categorizing}.
\item We propose a T-5 based architecture with a novel decoding method to generate a comparative sentence to underline the strengths of a particular item when specific features are important for a user. We evaluate our model automatically with Distinct-1/-2, Rouge-L, and BLEU scores and validate the results in a user study with human crowd workers and find that this method outperforms existing algorithms that were re-purposed for this task. 
\end{enumerate}
\section{Related Work}
\subsection{Natural Language Recommendation Generation}
Natural Language Generation (NLG) methods are a widely used procedure to create explanations for recommendations, and previous work uses different contextual information to train those language models. For example, \citet{dong2017learning, tang2016context} use item and user attributes (e.g. sentiment, user ID, rating) are used to generate reviews for specific users. \citet{ni2019justifying} use extracted aspects that distinguish user preferences and item features to generate justifications for recommendations. \citep{li2022personalized} use prompt learning to infuse context into pre-trained models for personalized recommendation justification. \citet{li2021personalized} train a personalized transformer for explainable recommendation. \citet{li2017neural} model item recommendation and explanation generation in a shared user and item embedding space, where they use predicted recommendation ratings as part of the initial state for explanation generations. \citet{lileigenerate} learn sentence templates from data and generate template-controlled sentences that comment about specific features. All these studies help to make recommendations more transparent and give an abstractive explanation for a recommendation to a user, but do not help to clarify which item features distinguish similar items, and do not show which aspects of an item make one item a better fit for a user over another.
\subsection{Comparisons as a Relative Evaluation Method}   
When every ranker in an information retrieval or recommendation task returns highly relevant items, recognizing meaningful differences between recommendations becomes increasingly difficult~\citep{10.1145/3477495.3531991}.
To address this issue, rather than viewing or explaining one item at a time, assessors can view items side-by-side and indicate the one that provides the better response to a query, hence comparing the items to allow fine-grained distinction~\citep{10.1145/3477495.3531991}. \citet{carterette2008here} show the effectiveness of comparisons in a study where they find that human assessors make pairwise preference judgments faster and more accurately. By considering two items side-by-side, assessors recognize fine distinctions between them that may be missed by pointwise judgments \citep{bah2015document, clarke2021assessing}. %Pairwise comparisons are also used to achieve fairness in recommendation through specifically ranking items in recommendation \citet{beutel2019fairness}. 
%Previous work has also identified that regular explanations cannot describe the differences between recommendations sufficiently \citep{yu2022towards}. 
\citet{yang2021comparative} claim that item recommendation in essence is a ranking problem to estimate a recommendation score for each item under a given user and that explaining relative comparisons among a set of ranked items from a recommender system is preferred \citep{yang2021comparative}. This leads to approaches which rank the items to maximize recommendation utility \citep{yang2021comparative, karatzoglou2013learning}. \citet{yu2022towards} propose an approach that jointly explains all documents in a search result list, such that differences in items can be incorporated. However, those comparisons are implicit due to the list-like nature of the representation and not explicitly communicated in natural language. In recommendation, pairwise ranking methods such as bayesian personalized ranking \citep{damak2021debiased} have previously been found to outperform pointwise models \citep{he2016vbpr}. These findings motivate our work. When it comes to providing recommendations or evaluations of products, the use of comparative sentences can be a valuable tool for providing clear and persuasive assessments. Comparative sentences allow the establishment of a frame of reference, clearly communicating the strengths and weaknesses of the item being evaluated in relation to others. This allows readers to better understand the relative merits of the recommendation and make informed decisions based on this information.
\subsection{Natural Language Comparisons}
To obtain natural language comparisons, \citet{zhang2010voice} identify comparative sentences in reviews and then assign sentiment orientations for recommendation. However, their work is extractive, and does not generate comparative sentences, hence experiences a cold-start problem and difficulties for personalization. \citet{yang2021comparative} mitigate this by providing a personalized extract-and-refine architecture. They select an existing sentence prototype that suits the desired comparison against a set of reference items, and refine it towards a recommendation explanation~\citep{yang2021comparative}. Their approach is different to our proposed method in that they do not provide an abstractive explicit natural language comparison, but an implicit comparison by showing explanations of different items. 
To enhance personalization of comparative explanations, methods from previous work for different use cases can be used. For example, previous work includes aspects of an item from its associated reviews for personalization~\citep{ai2018learning, he2015trirank, wang2018explainable, zhang2014explicit}.
%\citep{tan2021counterfactual} use aspects to create counterfactual explainable recommendation. 
\citet{le2021explainable} incorporate aspect-based comparative constraints into explainable recommendation models. This enables aspect-based comparisons of items with a template-based method to provide comparisons to a user. Our work contrasts this by going beyond extractive, template-based methods. Our work is generative, and uses aspects in the decoding method of our language model to increase personalization, compared to extractive or extract-and-refine methods~\citep{yang2021comparative, zhang2010voice}.
\section{Dataset Construction}\label{sec:comp_Sen}
\subsection{Comparative Sentence Extraction}\label{sec:comp_se}
Our goal is to generate natural language sentences that describe a comparison for an item. To compile a dataset of comparative sentences to train a Natural Language Generation (NLG) model, we need a large-scale corpus of comparative sentences. To be resourceful and avoid hand-labeling hundreds of thousands of review sentences (which may or may not be comparative), we collect a dataset large enough to train a comparative sentence classifier, and then classify sentences from reviews at scale with this classifier. This labeled dataset will then be used to train the NLG models. 

We pre-process subsets of the Amazon review corpus~\citep{ni2019justifying} according to a specific set of patterns. We first find potential comparative sentences by extracting sentences with a fixed set of commonly found words in comparative sentences with a regular expression. We search for common comparative words such as ``instead", ``better", ``worse", etc., and specific item descriptions that include numbers and letters delimited by a dash (e.g.~a template like ``FX-3200") present in the sentence. 
A randomly selected subset of the instances found by the regular expression was then manually labeled with crowds on Amazon Mechanical Turk as \textit{comparative} or \textit{non-comparative} sentences.  We require a 99\% approval rate and at leats 10,000 approved tasks for our crowd workers for this task. 
This labeled data was extended with comparative sentences from previous work, which are small-scale comparative sentence datasets~\citep{jindal2006identifying, panchenko2018categorizing, franzek2018categorization}. Our total dataset for training the comparative sentence classifier consists of 7,819 samples, which is then used to train the classifier to find comparative sentences in large corpora of reviews.  

\begin{figure*}
    \centering
   \includegraphics[width=\textwidth]{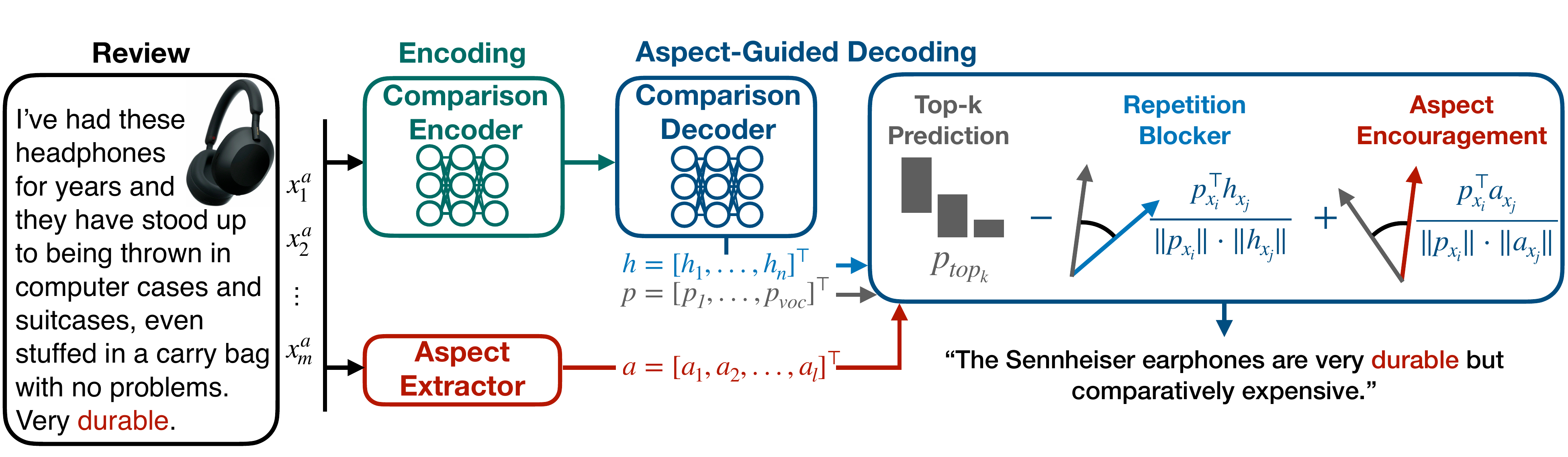}
    \caption{We train an encoder-decoder architecture to generate comparative sentences. To guide the decoding process toward relevant item features, we use a novel aspect encouragement term between the item aspects extracted from all reviews for an item and the next token to be generated in the decoding process. This increases the probability of including aspects into the decoding. Once an aspect is mentioned, the cosine similarity between the generation and the next token avoids repetitions~\citep{su2022contrastive}.}
    \label{fig:pipelinenew}
\end{figure*}
\subsection{Automatically Labeling Comparative Sentences} 
To compile a larger corpus of comparative sentences, we automatically label instances from a subset of Amazon Musical Instruments and Amazon Electronics to extract comparative sentences for each item. To reliably find comparative sentences automatically, we use our previously manually labeled dataset to train a BERT classification model. 
We tokenize our data instances with the BERT tokenization scheme~\citep{devlin2018bert} and pad sentences to a fixed length. The model is then fine-tuned for 4 epochs with a linear classification layer to predict comparative sentences. The comparative sentence classification training data was split into training/validation and test sets and the best model was chosen on the validation set. We provide comparative sentences in our dataset which were classified with our BERT classifier reaching an F-1 Score of $89\%$. In our final large-scale labeled dataset, we use instances that are classified as comparative by the classifier with prediction confidence of $>90\%$.\footnote{This probability is indicated by the softmax of the final linear classification layer.} The comparative sentences will be further used as a ground-truth for our natural language generation procedure. Statistics of our datasets are shown in Table \ref{tab:statistics}.
\subsection{Obtaining Item Aspects}
To infuse control into our language generation task and personalize  generation toward item features that are relevant to a particular user, we obtain aspects from all reviews of an item following the method from \citet{zhang2014users}. We build a sentiment lexicon to extract fine-grained aspects from the reviews of the dataset and their associated sentiments. For example, a review like \emph{``I like the sound of these headphones"} would have a positive sentiment towards the aspect \emph{``sound"}. We extract both positive and negative aspects in our dataset, but proceed to use only aspects extracted with a positive sentiment to feed into our language generation pipeline. 
%We decided to focus on using only positive sentiments because these occurred more frequently in the review corpus. However, our methods can be used in a similar fashion with negative sentiment aspects.
To summarize, our dataset consists of item reviews, item aspects and their sentiments, comparative sentences for each item, as well as product and user information.
\section{Comparative Sentence Generation}
\subsection{Personalized Aspect-Guided Generation (AGG)} 
To generate comparative sentences, we first use a T5-small transformer~\citep{raffel2019exploring} model, which we fine-tune to generate comparative sentences about an item given a user review. 
The T5 transformer is most commonly used for extracting information for summarization tasks. The idea is that this model, when fine-tuned, is suitable to generate comparative sentences while extracting the `gist' of the product information from the user review. The transformer consists of an encoder-decoder architecture that takes an existing review of the product as input and is fine-tuned to output a comparative sentence. The model is trained for each dataset separately for 9 epochs using cross-entropy loss. Stochastic decoding and beam search typically optimize for generating high likelihood sequences rather than diverse ones~\citep{ippolito2019comparison}. We hence extend our model capabilities by a novel decoding method, which can, but is not forced to, include one or more aspects of each item that can be personalized to each user (hence \emph{personalized, aspect-guided decoding}). At each decoding step, the idea is to select the next token from the set of the most probable candidates predicted by the model while encouraging the model to mention certain relevant aspects of the product, that stand out compared to a different product. 
We define a set of tokens $V^k$ to be the set of top-k predictions from the model's probability distribution $p_\theta(\cdot|x_{1:t-1}) $, and extend this set by the relevant aspects for the item. Given the previous context $x_{1:t-1}$ at time step $t$, and relevant aspect tokens $\{a_1, ..., a_n\}$, the selection of the output $x_t$ is computed as a combination of the model confidence, degeneration penalty~\citep{su2022contrastive}, and aspect encouragement:
\begin{equation}
\begin{split}
x_t = \underset{v \in V^{(k)}}{argmax} \Bigl\{ (1- \alpha - \beta) \times p_\theta(v|x_{1:t-1})  \\
- \alpha \times max\{s(h_v, h_{x_j}): 1 \leq j \leq t-1\} \\
+ \beta \times max\{s(h_v, h_{a_i}): 1 \leq i \leq n\} \Bigl\}
\end{split}
\end{equation}\label{dec_eq}
Where the model confidence is defined as $p_\theta(v|x_{1:t-1})$. The novel aspect encouragement term $max\{s(h_v, h_{a_i}): 1 \leq i \leq n\}$ is defined as the cosine similarity $s = \frac{h_{x_i}^\top h_{x_j}}{\lVert h_{x_i} \rVert \cdot \lVert h_{x_j} \rVert}$ between the next token and the aspect tokens. The degeneration penalty $max\{s(h_v, h_{x_j}): 1 \leq j \leq t-1\}$ is defined as the cosine similarity between the next token and the previous context. 

By definition, the cosine similarity $s(h_{x_j}, h_{x_j})$ of the identical token $x_j$ is 1.0. In Equation \ref{dec_eq}, the first term, model confidence, is the probability of candidate $v$ predicted by the model. The second term, degeneration penalty, measures how discriminative is candidate $v$ with respect to the previous context $x_{<t}$. Specifically, the discriminative power is defined as the maximum cosine similarity between the representation of $v$ and that of all tokens in $x_{<t}$. Here, the candidate representation $h_v$ is computed by the model given the concatenation of $x_{<t}$ and $v$. A larger degeneration penalty of $v$ means it is more similar to the previous context, therefore more likely leading to model degeneration~\citep{su2022contrastive}. 
This is an important factor for our aspect encouragement term. Intuitively, once an aspect is included in the generation, the cosine similarity between the aspects and the next token will rise for similar aspects, making it less likely to be mentioned again.  The hyperparameter $\alpha \in [0, 1]$ regulates the importance of the degeneration term, $\beta \in [0, 1]$ regulates the importance of the aspect encouragement term. When $\alpha = 0, \beta = 0$, this procedure becomes the greedy search method. 

The unique advantages of our method are: (i) the cosine similarity encourages the exact word, but also similarly embedded words, that might fit well to the product; (ii) it is possible to encourage multiple words and we do not need to stop after one word is present in the generated sentence; (iii) the encouragement of specific words is automatically restricted by the degeneration property \citep{su2022contrastive}, which discourages repetitiveness; and (iv) if the context and aspects are too dissimilar, an aspect is not forced to be included during generation.

\begin{table}
    \centering
    \resizebox{0.48\textwidth}{!}{
    \begin{tabular}{l|rrrr}
\toprule
Dataset &  Train &  Val &  Test &  \#Items \\
\midrule
    Musical Instruments & 131,714 & 2,835 &  2,381 &   19,133 \\
     Electronics & 118,563 & 2,050 &  2,273 &    7,123 \\
\bottomrule
\end{tabular}}
    \caption{Statistics of the comparative review data from Musical Instruments (MI) and Electronics (E) with the number of unique items.}
    \label{tab:statistics}
\end{table}

\begin{table*}[t]
    \centering
    \resizebox{\textwidth}{!}{
    \begin{tabular}{llllllll|lllllll}
    \toprule
    \multicolumn{1}{l}{Dataset} & \multicolumn{7}{l}{Musical Instruments}& \multicolumn{5}{l}{Electronics} & \\
    \midrule
         Model &  D-1 &  D-2  &  B-1 & B-2 & RL-P & \% Comp. & \% Asp. & D-1 &  D-2  
         & B-1 & B-2 & RL-P & \% Comp. & \% Asp.\\
\midrule
Justif-Extract~\citep{ni2019justifying} & 0.487               & 0.857          & 0.566          & 0.130          & 0.458          & 0.077          & 0.790          & 0.466          & 0.841          & 0.490          & 0.074          & 0.365          & 0.112          & 0.670          \\
Comp-Extract (ours)             & 0.276               & 0.722          & 0.540          & 0.198          & 0.491          & 1.000          & 0.691          & 0.267          & 0.721          & 0.443          & 0.078          & 0.413          & 1.000          & 0.736          \\
\midrule
T5-PPLM-BoW                     & 0.082               & 0.394          & 0.464          & 0.155          & 0.282          & 0.925          & 0.741          & 0.105          & 0.449          & 0.423          & 0.164          & 0.312          & \textbf{0.823} & 0.803          \\
T5-PPLM-BoW-Discr               & 0.089               & 0.411          & 0.476          & 0.162          & 0.297          & \textbf{0.954} & 0.767          & 0.212          & 0.616          & 0.364          & 0.122          & 0.233          & 0.433          & 0.619          \\
T5-Stochastic (ours)            & 0.299               & \textbf{0.768} & 0.481          & 0.077          & 0.423          & 0.783          & 0.647          & 0.313          & 0.773          & 0.481          & 0.126          & 0.447          & 0.718          & 0.851          \\
T5-AGG (ours)                   & \textbf{0.312}      & 0.775          & \textbf{0.594} & \textbf{0.163} & \textbf{0.516} & 0.884          & \textbf{0.768} & \textbf{0.321} & \textbf{0.779} & \textbf{0.605} & \textbf{0.269} & \textbf{0.560} & 0.765          & \textbf{0.927} \\
        
\bottomrule
    \end{tabular}}
    \caption{Automatic evaluation metrics for our trained comparative language models (bottom), compared to extractive methods (top). Our methods outperform baselines in diversity, Rouge-L Precision, BLEU score and aspect awareness.}
    \label{tab:autom}
\end{table*}

\subsection{Baseline}
We re-purpose the method of PPLM, previously used for topic control or language detoxification, for comparative sentence generation~\citep{Dathathri2020Plug}. PPLM based language models control the language model at every step of generation, towards higher log-likelihood of a conditional model $P_\theta(a|x_{1:T})$ and an unmodified language model $P_\theta (x_{1:T})$.  In the first baseline, we use the fine-tuned T5 generation model and feed item aspects $a$ as a Bag-of-Words into the model (PPLM-BoW) and update the gradient direction to include the words (attributes)~\citep{Dathathri2020Plug}. The log likelihood of the attribute model is given by $ log(P_\theta(a|x_{1:T}) = log(\sum_i^n P_\theta(x_{t+1})[a_i])$.
To extend the capabilities of the first baseline, we include our pre-trained comparative sentence classifier. With this discriminator model $f$, the discriminative attribute model objective is given by $log(P_\theta(a|x_{1:T}) =  log(f(o_{:t+1}, o_{:t+2}))$ for logit vector $o_{:t+1}$ in step $t+1$, which encourages comparativeness of generated sentences inside the PPLM procedure (PPLM-BoW-Disc) in addition to the attribute model. This procedure alters the output token probabilities if the currently generated sentence is not determined to be a comparative sentence by the classifier.
For PPLM, we use a stepsize of 0.03 and sample with geometric mean fusion scale set to 0.9, and the Kullback-Leibler divergence scale set to 0.01. We consider a gradient length of 30 and 4 gradient adaptation iterations. The maximum length for all generated samples is 350. All other parameters were kept equal to~\citet{Dathathri2020Plug}.
We additionally compare to different recommendation explanation methods, that are based on extractive reviews. We compare our method to comparative sentences that were written in previous reviews, that can be obtained with our comparison classifier, as well as extracted recommendation justifications (using \citep{ni2019justifying}). This gives us an indication on if the additional personalization and abstraction capabilities of our method are helpful, compared to purely extractive approaches.
\subsection{Evaluation} 
\subsubsection{Automatic Evaluation.}
To evaluate the \emph{diversity} of the generated text automatically we use average Distinct-1 (D-1) and Distinct-2 (D-2)~\citep{li2015diversity} metrics to calculate the ratio of unique 1- and 2-grams with respect to all generated tokens. To automatically evaluate \emph{fidelity} of the generated sentences, we use Rouge-L~\citep{lin2004rouge} to measure overlapping n-grams between the source input including aspects, and the generated comparative output tokens. Rouge-L has the advantage that it does not require consecutive matches, but in-sequence matches that reflect sentence level word order, and it automatically includes longest in-sequence common n-grams.  We additionally evaluate on the commonly used BLEU-1 (B-1) and BLEU-2 (B-2) score, which is based on the precision of n-gram matches between the generated sentence and reference~\citep{papineni2002bleu}.
%Rouge-L precision (RL-P) explains the percentage of n-grams in the reference that are present in the generated sentence, while recall (RL-R) explains the percentage of n-grams in the generated sentence that are present in the reference. 
We evaluate the \emph{comparativeness} of the generated samples using predictions of our classifier from Section \ref{sec:comp_se} (\% Comp.). We evaluate aspect awareness by percentage of generated sentences that contain at least one of the item aspects (\% Asp.). 
\subsubsection{Human Evaluation}
To further evaluate our method, we conduct human experiments on (1) \emph{Comparativeness}, where we ask human raters if a sentence is comparing the item or a feature of the item; (2) \emph{Relevance}, where we ask if the generated sentence is relevant to the item(s) in question, and (3) \emph{Fidelity} to see if the generated sentence is truthful with respect to the review and aspect information present.
We sample 100 generated examples for the best performing model and decoding combinations for baselines and our models. Human annotators are asked to give an agreement/disagreement score. Each example is rated by at least three annotators. We combine results by a majority vote to filter for low-quality raters and then calculate the accuracy on the results. 
\section{Results}
We provide a dataset of $258,816$ comparative sentences, associated reviews and user information. 
%For Musical Instruments we provide $131,714$ comparative sentences in the training set, $2,835$ in the validation set and $2,381$ in the test set for $19,133$ items. For Electronics we provide $118,563$ comparative sentences in the training set, $2,050$ in the validation set and $2,273$ in the test set for $7,123$ items.
We validate our automatic labeling approach in a human evaluation study, where we randomly select 200 comparative sentence instances to be labeled by three different crowd workers. For 92\% of instances, at least two out of three crowd workers agreed that the sentence is a comparative sentence. 

In our automatic evaluation (Table \ref{tab:autom}), we observe that a large fraction of comparative sentences contain at least one item aspect (69-74\%), especially when noting that those aspects were independently extracted by methods from previous work \citep{zhang2014users}. This supports the hypothesis that comparative sentences underline important features of an item. We observe that our methods show high D-1 and D-2 metrics, suggesting that the generated content is more diverse compared to the PPLM baseline. 
%Additionally, the metrics appear to be similar to the extractive approach, but with increased aspect percentage for our aspect-guided generation method. 
Overall, our methods outperform the baseline in 6 out of 7 tasks, shown in Table \ref{tab:autom}. 

For human evaluation results (\% of the generated sentences that human crowdworkers agree on), the baseline T5-PPLM-BoW-Disc achieves 73.5\% for comparativeness, 79\% for relevance and 77.5\% for fidelity. Our T5-AGG achieves 86.5\% for fidelity, 92\% for relevance and 69.5\% for comparativeness. For human evaluation, we see that our aspect guided generation of comparative sentences achieves higher fidelity and relevance scores compared to the best performing PPLM model in the automatic evaluation tasks, whereas the aspect-guided decoding model shows a smaller percentage of comparative sentences in human evaluation. Overall, the results are aligned with automatic metrics. Compared to the PPLM baseline, our method is more computationally efficient, requiring only one forward pass through the model, whereas PPLM requires an additional forward pass through the Bag-of-Words or discriminator model, as well as an additional backward pass to update the internal latent representations of the language model.
\section{Limitations}
One limitations for the task of comparative sentence generation for product recommendation is the lack of a standardized evaluation metric. We can use automated metrics such as BLEU \citep{papineni2002bleu} and distinct \citep{li2015diversity}, but we have to be mindful that these metrics were developed for different purposes (e.g. translation tasks). Our work extends automatic evaluation with human evaluation on comparativeness, relevance and fidelity for this purpose. 

While large language models \citep{ouyang2022training} are powerful, they may not provide personalized behavior. Our comparative sentence generation method is designed to be personalized by incorporating user aspects into the generation procedure, and our method is based on the assumption that it is possible to obtain these kinds of user information.  Compared to large language models, our study provides a customized language models tailored to a specific domain. Large language models are typically general-purpose models that have been trained on a wide range of text data. For some tasks or domains, like specifically generating comparative sentences, it may be beneficial to have a language model that is trained on a specific type of data to be more accurate and relevant for this specific task. Additionally, large language models may raise concerns about privacy, as they require large amounts of data to be trained effectively. In some cases, users or companies may prefer to use a language model that is trained on their own data, rather than sharing their data with a large language model provider. Our method can be trained in few epochs on small amounts of data, while still providing truthful comparative sentences. 

\section{Conclusion}
This work studies the problem of generating comparative sentences for item recommendations. We provide a pipeline to extract comparisons from large review corpora and a large-scale annotated comparative sentence dataset, that is $35\times$ larger than previous datasets. Our work indicates that it is possible to abstractively generate fluent comparative sentences that highlight relevant item features. Compared to previous extractive or extract-and-refine methods, our generated comparative sentences can be personalized with a novel aspect aware decoding method, that outperforms the baseline methods. Compared to baseline generative alternatives, our model shows increased diversity and relevance when evaluated with automatic and human evaluation metrics.

%test
{
\small
\bibliography{emnlp2022}
\bibliographystyle{acl_natbib}
}
\end{document}